\documentclass[10pt,twocolumn,letterpaper]{article}

\usepackage{cvpr}              

\usepackage{kotex}

\usepackage{array}
\usepackage{booktabs}
\usepackage[table]{xcolor}

\definecolor{cvprblue}{rgb}{0.21,0.49,0.74}
\usepackage[pagebackref,breaklinks,colorlinks,allcolors=cvprblue]{hyperref}

\def\confName{CVPR}
\def\confYear{2027}

\title{PSC-Gen: Physical Sequence Conditioning for Video Generation \\ via Agentic Visual Planning}

\author{
Wonsik Shin$^{1}$ \quad
Jinju Jang$^{1}$ \quad
Junhoo Lee$^{2}$ \quad
Yeonjin Chang$^{1}$ \quad
Jimyeong Kim$^{1}$ \quad
Nojun Kwak$^{1}$\\[4pt]
$^{1}$Seoul National University \qquad
$^{2}$Korea Advanced Institute of Science \& Technology\\
{\tt\small \{wonsikshin, jinju.jang, yjean8315, wlaud1001, nojunk\}@snu.ac.kr} \qquad
{\tt\small junhoo.lee@kaist.ac.kr}
}

\makeatletter
\apptocmd{\@maketitle}{%
    \noindent
    \begin{minipage}{\textwidth}
        \centering
        \vspace{-2em}
        \includegraphics[width=\linewidth]{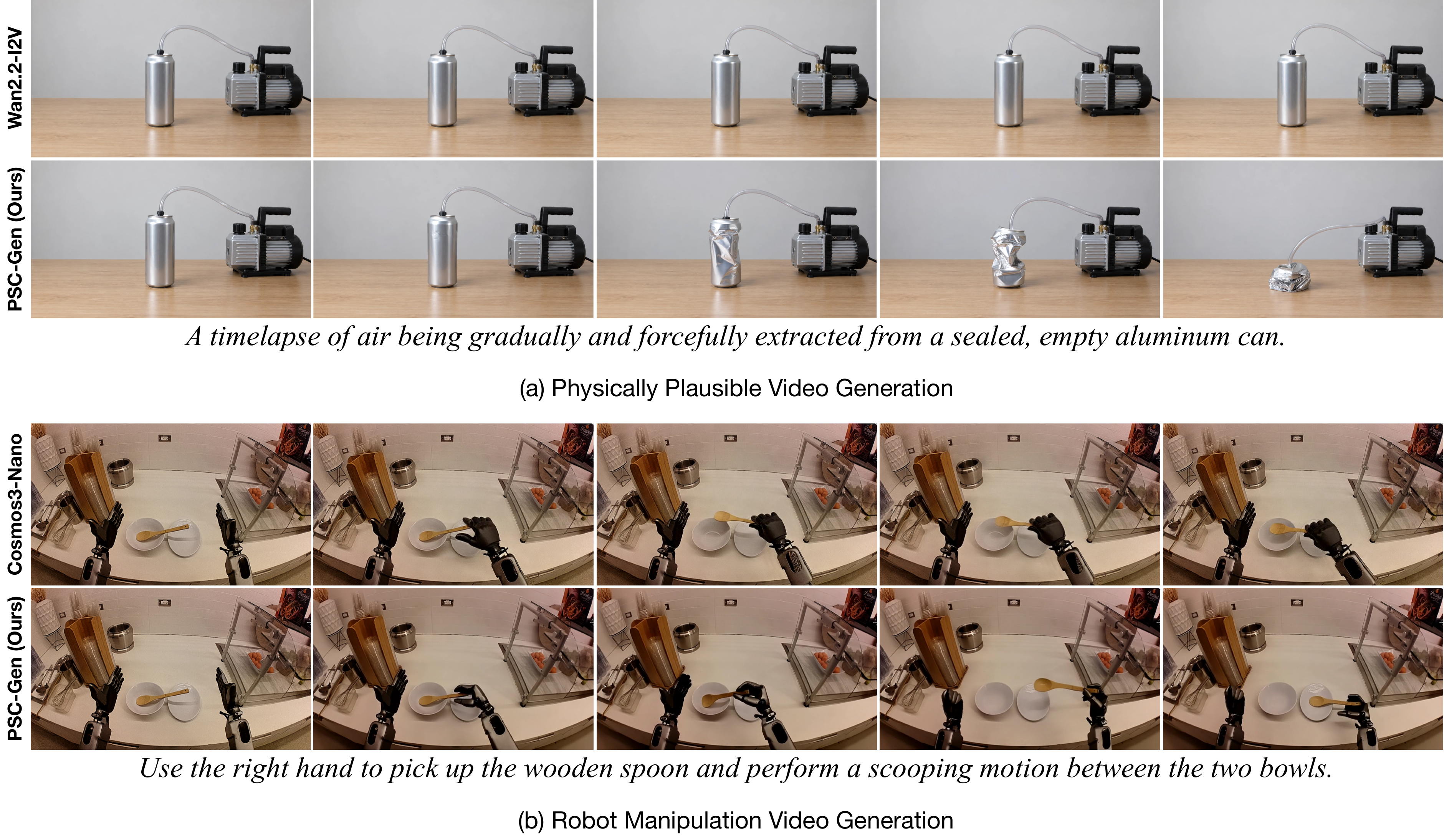}
        \vspace{-1.3em}
        \captionof{figure}{
        \textbf{Comparison between baseline models and PSC-Gen in (a) Physically Plausible Video Generation and (b) Robot Manipulation Video Generation.} In (a), Wan2.2 fails to reason from the text prompt and remains nearly static throughout the video, whereas PSC-Gen depicts the can crushing gradually in a physically natural manner. In (b), when instructed to scoop between two bowls, the baseline repeatedly scoops within the left bowl, while PSC-Gen accurately scoops from the left bowl into the right bowl.
        }
        \label{fig:teaser}
    \end{minipage}
    \par\vspace{1em}
}{}{}
\makeatother

\begin{document}
\maketitle
\begin{abstract}

Recent advances in video diffusion models have enabled the synthesis of high-fidelity, visually compelling videos.
However, despite being trained on internet-scale video data, understanding underlying physical laws and generating physically plausible dynamics remain challenging.
To mitigate this, conditioning video generators on explicit physical reasoning has emerged as a promising direction.
Nevertheless, existing conditioning approaches often fail to incorporate detailed physical guidance into generation, resulting in physically implausible motion or visual artifacts.
We propose PSC-Gen, a training-free framework that employs an agentic approach to synthesize visual conditions and inject their core dynamics via low-frequency filtering.
By iteratively planning physical states and verifying generated keyframes, our Physics Agent constructs visually and physically consistent keyframe sequences.
We then interpolate these keyframes into a continuous physical sequence and inject only its low-frequency components, steering the video model to preserve physical dynamics while synthesizing fine visual details.
Qualitative and quantitative evaluations across physics-informed and robotic manipulation benchmarks demonstrate that PSC-Gen faithfully follows instructions and generates physically plausible videos.

\end{abstract}
\vspace{-3em}

\section{Introduction}
\label{sec:intro}

Video diffusion models have advanced rapidly, enabling the synthesis of high-fidelity scenes with natural appearance and coherent motion~\cite{wan22,videodiffusionmodels2022ho}.
Beyond content creation, they are increasingly adopted as world models for predicting future scenes and simulating real-world dynamics~\cite{jang2025dreamgen,nvidia2026cosmos3omnimodalworld, unisim}.
These models are expected to faithfully capture real-world physics while maintaining the semantics of the input prompts.
Achieving this requires modeling continuous spatiotemporal dynamics, including the physical consequences of the initial state and the precise timing of events.
However, synthesizing such complex physical interactions demands far more than realistic rendering.
Bridging this gap between visual realism and physical plausibility stands as a key open problem in video generation.

Recent works have addressed this challenge by incorporating physical priors during model training~\cite{wang2025wisa, narayanan2026phyco, pathak2026physvid}, or by grounding the generation process in physically informed conditioning signals~\cite{xue2025phyt2vllmguidediterativeselfrefinement, yang2025vlipp, wang2026chainCoECT}.
While conditioning approaches bypass extensive data collection and costly retraining, their effectiveness depends critically on the quality of the conditioning signals.
Existing methods either use Large Language Models (LLMs)~\cite{openai2024o1preview, openai2025gptoss120bgptoss20bmodel} to expand brief prompts into detailed descriptions~\cite{xue2025phyt2vllmguidediterativeselfrefinement} or leverage Vision Language Models (VLMs)~\cite{openai2024gpt4ocard} to provide explicit visual cues such as bounding box trajectories~\cite{yang2025vlipp} and keyframe images~\cite{wang2026chainCoECT}. 
Although visual conditioning offers stronger grounding, vague signals omit crucial physical details, whereas overly strict conditions often introduce visual artifacts~\cite{wang2026chainCoECT}. 
Effectively injecting rich visual information without propagating artifacts from these conditions remains a critical yet largely underexplored problem.

To mitigate this issue, we propose \textbf{PSC-Gen} (Physical Sequence Conditioning for Video Generation), a training-free framework that guides video synthesis through a physically grounded visual sequence. 
To construct an accurate conditioning sequence, we employ a Vision Language Model (VLM) as a Physics Agent.
Although the Physics Agent formulates a coherent physical plan, the inherent stochasticity of image generation models can still yield keyframes that fail to reflect the planned physical states.
To ensure physical consistency across keyframes, the agent inspects each synthesized image against physical principles and iteratively regenerates implausible candidates until every keyframe passes verification.
To provide dense spatiotemporal guidance across frames, we then use an efficient video interpolation model to convert these verified keyframes into a continuous physical sequence.
Finally, we compute the guidance velocity that directs the noisy latent toward this physical sequence and use it to steer the predicted denoising velocity during diffusion sampling.
However, because directly conditioning the raw guidance velocity can propagate visual artifacts, we extract only its low-frequency component and inject it into the predicted velocity.
With this design, PSC-Gen effectively instills physical dynamics into the generation process while allowing the diffusion model to synthesize high-frequency visual details.

We validate PSC-Gen across a broad suite of benchmarks spanning general physical dynamics and robotic manipulation. 
Our method consistently outperforms existing baselines, achieving superior physical realism on PhyGenBench~\cite{phygenbench} and Physics-IQ~\cite{physicsiq} while demonstrating strong generalization to embodied world modeling on DreamGenBench~\cite{jang2025dreamgen}. 
Furthermore, qualitative comparisons confirm that our approach generates visually compelling and physically plausible videos, and extensive ablation studies verify the effectiveness of both the Agentic Visual Planning and Physical Sequence Conditioning.

Our main contributions are as follows:
\begin{itemize}
    \item We introduce \textbf{PSC-Gen}, a training-free video generation framework that delivers rich physical guidance without propagating visual artifacts from the conditioning signals.
    \item We propose \textit{Agentic Visual Planning} and \textit{Physical Sequence Conditioning} to construct physically consistent visual sequences and steer the diffusion denoising process toward faithful spatiotemporal dynamics.
    \item We demonstrate across multiple physics and world modeling benchmarks that our approach consistently surpasses baseline models in both physical fidelity and prompt adherence.
\end{itemize}

\begin{figure*}[t]
    \centering
    \includegraphics[width=\textwidth]{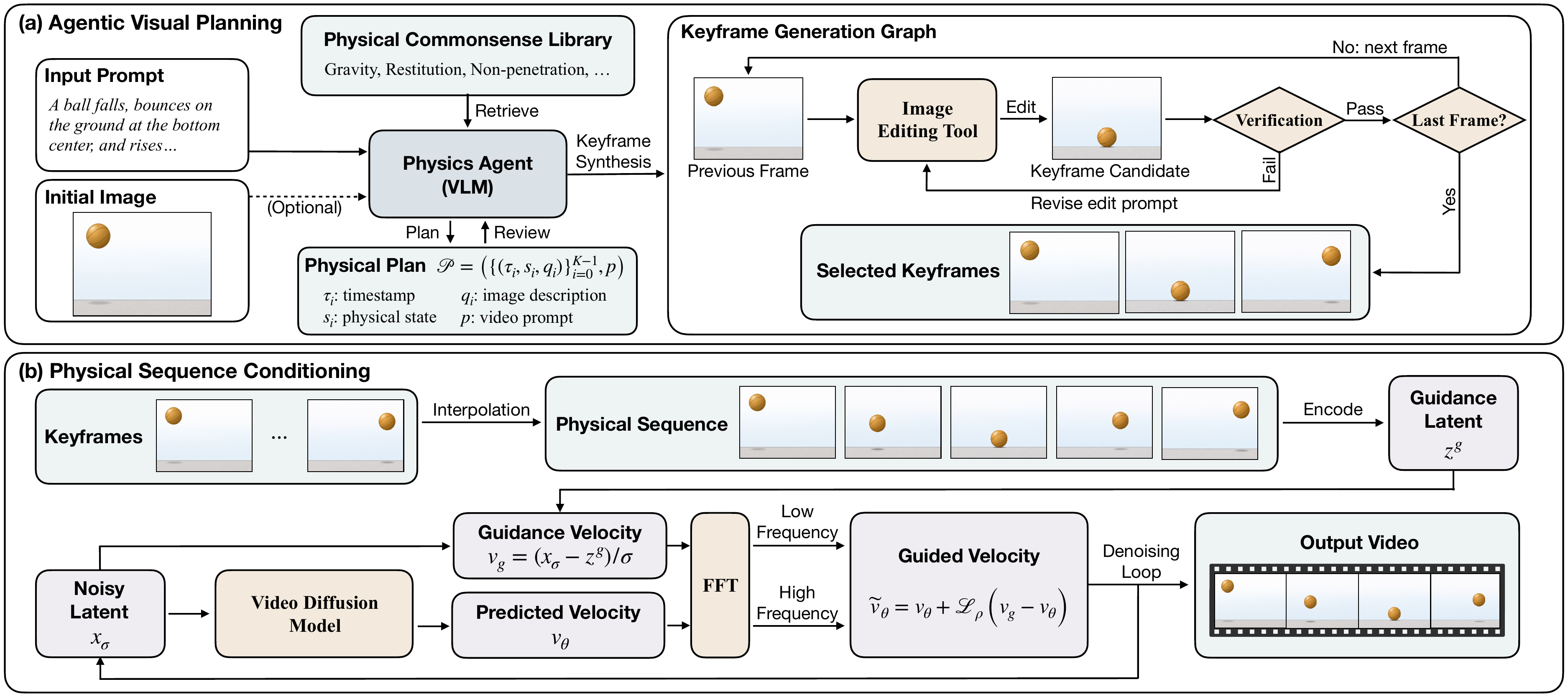}
    \vspace{-1.5em}
    \caption{
    \textbf{Method Overview.} PSC-Gen generates physically plausible videos through (a) \textit{Agentic Visual Planning} and (b) \textit{Physical Sequence Conditioning}. A vision-language model (VLM) acts as a Physics Agent to plan physical states and iteratively generate and verify keyframes. These keyframes are interpolated into a physical sequence, which guides diffusion sampling through low-frequency velocity injection. This design effectively transfers physical dynamics while allowing the diffusion model to synthesize fine visual details.
    }
    \label{fig:method_overview}
    \vspace{-0.7em}
\end{figure*}

\section{Related Work}
\label{sec:relatedwork}

Synthesizing videos that faithfully follow physical laws remains a core challenge in video generation. 
Training-based approaches~\cite{wang2025wisa,pathak2026physvid,narayanan2026phyco,videorepa,physhpo,Yuan_2025_NewtonGen} address this by finetuning diffusion models on extensive physical datasets, yet they require costly optimization~\cite{wang2025wisa}.
In contrast, conditioning-based approaches~\cite{xue2025phyt2vllmguidediterativeselfrefinement,yang2025vlipp,wang2026chainCoECT,PhysAnimator,psivg,physgen} bypass training by guiding synthesis through external signals.
Textual refinement methods such as PhyT2V~\cite{xue2025phyt2vllmguidediterativeselfrefinement} iteratively expand prompts via video feedback, but incur heavy inference overhead and struggle to convey precise spatial geometry through text alone.
To provide explicit spatial grounding, VLIPP~\cite{yang2025vlipp} converts VLM-queried motion trajectories into optical flow and structured noise, yet such coarse abstractions cannot fully capture complex physical dynamics.
At the image level, CoECT~\cite{wang2026chainCoECT} synthesizes keyframes from causal events and uses them as starting points for denoising.
However, keyframes generated in a single pass often violate physical constraints, and directly conditioning on them propagates visual artifacts into the synthesized video.
To overcome these limitations, PSC-Gen combines Agentic Visual Planning and Physical Sequence Conditioning to construct verified physical sequences and selectively inject their low-frequency dynamics into the generation process.

\section{Method}
\label{sec:method}

We propose \textbf{PSC-Gen} (\textbf{P}hysical \textbf{S}equence \textbf{C}onditioning for Video \textbf{Gen}eration), a training-free framework that generates physically plausible videos faithful to input prompts (\Cref{fig:method_overview}).
Rather than relying on a pretrained video diffusion model to infer complex physics directly from text, PSC-Gen bridges explicit physical reasoning with generative denoising.
Specifically, it translates a text prompt into a grounded physical sequence to guide low-frequency global dynamics, while leaving high-frequency visual details to be synthesized by the diffusion model itself.

\subsection{Agentic Visual Planning}
\label{sec:agentic_visual_planning}

\paragraph{Video Planning.}
As a short prompt leaves the intermediate progression ambiguous~\cite{wu2025mint}, we first employ a Vision Language Model (VLM)~\cite{qwen38} as a \emph{Physics Agent}~\cite{yao2023react} to convert the input prompt into a structured plan for video generation.
From the prompt, the agent identifies the initial scene entities and defines the constraints governing their interactions.
When an initial image $I_0$ is provided alongside the prompt, the agent uses it to ground the initial state.
To equip the agent with explicit physical knowledge, we construct a physical commonsense library containing 28 knowledge cards across 8 domains, such as fluids and mechanics.
Each card provides fundamental principles (e.g., ``a collision transfers momentum between interacting bodies'') and governing formulas.
By retrieving the relevant cards, the agent formulates concrete constraints and motion plans.

Based on this grounded context, the Physics Agent constructs a sequence of physical anchors that capture critical state transitions, such as contact and impact, and selects the most informative anchors as key timings for keyframe generation.
This adaptive selection assigns dense timings to intervals with complex dynamics and sparse timings to less dynamic scenes, avoiding the redundancy of uniform temporal sampling.
Formally, the resulting plan for video generation is represented as
\begin{equation}
    \mathcal{P}
    =
    \left(
         \{(\tau_i,s_i,q_i)\}_{i=0}^{K-1},
        p
    \right),
    \label{eq:physical_plan}
\end{equation}
where $K$ is the number of target keyframes (typically 3 to 6 selected anchors).
For each timing, $\tau_i$ denotes the timestamp in seconds, $s_i$ specifies the physical state describing each object's position and status, and $q_i$ is the text prompt for synthesizing the corresponding keyframe, while the video prompt $p$ summarizes the complete plan to guide final video generation.
To maintain consistency across frames, the initial state $s_0$ specifies all scene entities, whereas subsequent states $s_i$ describe only state changes, inheriting unchanged properties from preceding states.
These states serve as the reference for keyframe verification.
Finally, the Physics Agent inspects $\mathcal{P}$ for physical or logical inconsistencies and revises the plan with corrective feedback if any issues are detected.
The detailed video planning procedure is provided in the supplementary material.

\paragraph{Keyframe Generation.}
Even with an accurate plan, synthesizing faithful keyframes remains challenging due to the inherent stochasticity of image generation models~\cite{li2025enhancing}, which can produce varying outputs from the same prompt.
We therefore employ the Physics Agent to iteratively generate and verify keyframes until they accurately reflect the target states in the verified plan $\mathcal{P}$.
When an initial image $I_{0}$ is provided, it serves directly as the first keyframe; otherwise, $I_{0}$ is synthesized from the initial prompt $q_0$.
Starting from $I_{0}$, we sequentially generate each subsequent keyframe $I_{i}$ according to the prompt $q_i$ by editing the preceding keyframe $I_{i-1}$.

To verify each synthesized output, the agent visually inspects the candidate keyframe against the target state $s_i$ and the preceding keyframe $I_{i-1}$.
Specifically, the agent verifies whether the intended state transition is clearly realized and whether unrelated scene properties remain intact.
When a candidate fails verification, the agent diagnoses the specific visual discrepancy from the rejected image and autonomously reformulates the editing prompt $q_i$ to correct the error in the next attempt.
Crucially, every retry edits the same reference keyframe $I_{i-1}$, using the rejected candidate solely as diagnostic context.
Throughout this reflection loop, the target state and evaluation criteria remain strictly fixed, ensuring that prompt revisions sharpen the instruction without relaxing physical requirements.
Once a candidate passes inspection, the agent commits it as the keyframe $I_i$ and advances to the next transition until the complete keyframe set is obtained.

\subsection{Physical Sequence Conditioning}
\label{sec:physical_sequence_conditioning}

\paragraph{Physical Sequence Synthesis.}
Although the verified keyframes capture critical physical states, their temporal sparsity provides insufficient guidance for the video denoising process~\cite{zhu2025generative}.
Bridging these discrete states requires dense intermediate transitions.
However, simple temporal heuristics distort the underlying motion.
Duplicating keyframes across intermediate timestamps reduces the sequence to piecewise static scenes with abrupt jumps~\cite{Zhang_2025_CVPR_Continuous}, whereas linear interpolation blends pixel intensities, causing moving objects to gradually vanish at their initial positions and reappear at their target locations rather than following a continuous path~\cite{wang2025generative}.
To establish plausible trajectories between keyframes, we instead employ a video diffusion model that interpolates keyframes into a temporally dense physical sequence.
Specifically, we adopt a distilled video model~\cite{hacohen2025ltx25} that synthesizes a complete sequence from multiple image conditions in under 60 seconds, introducing negligible computational overhead compared to the primary video diffusion backbone~\cite{wan22}, which requires approximately 30 minutes per video.

\paragraph{Noisy Sequence Conditioning.}
The synthesized physical sequence establishes a plausible motion trajectory from the verified plan and keyframes, but its visual fidelity is constrained by the limited capacity of the fast interpolation model.
Directly conditioning the primary video generator on this imperfect sequence would propagate visual artifacts and degrade the final output.
However, despite its fine-grained imperfections, the sequence still captures the verified keyframe states and coarse temporal dynamics~\cite{yang2025vlipp}.
Because low-frequency components of diffusion latents govern global spatial layout and coarse motion whereas high-frequency components encode fine visual details~\cite{wu2023freeinit,ilvr}, 
we condition the denoising process exclusively on the low-frequency spatial components of the physical sequence and delegate high-frequency detail synthesis to the more capable primary video diffusion backbone.

Let $\mathcal{F}_{xy}$ denote the two-dimensional Fourier transform over the spatial axes of a video latent, applied independently to each channel and frame.
We define the spatial low-pass operator as
\begin{equation}
    \mathcal{L}_{\rho}(u)
    =
    \mathcal{F}_{xy}^{-1}
    \left(
        M_{\rho}\odot\mathcal{F}_{xy}(u)
    \right),
    \label{eq:spatial_lowpass}
\end{equation}
where $\odot$ denotes element-wise multiplication and $M_{\rho}$ is a hard radial mask defined by
\begin{equation}
    M_{\rho}(\nu_x,\nu_y)
    =
    \mathbf{1}
    \left[
        \nu_x^2+\nu_y^2\leq\rho^2
    \right].
    \label{eq:frequency_mask}
\end{equation}
Here, $\nu_x$ and $\nu_y$ denote spatial frequencies normalized by the Nyquist frequency along each respective axis, and $\rho$ controls the retained bandwidth.
Restricting this operation to the spatial axes removes high-frequency artifacts while leaving the temporal dimension untouched to preserve continuous motion.

To inject this low-frequency signal into the primary video diffusion model, we first encode the physical sequence $V_{\mathrm{g}}$ using the diffusion model's video encoder to obtain the guidance latent $z^{\mathrm{g}}$.
Let $x_{\sigma}$ denote the noisy video latent at noise level $\sigma\in[0,1]$, with sampling proceeding from high to low noise.
Under flow matching~\cite{lipman2023flow}, the noisy latent is expressed as
\begin{equation}
    x_{\sigma}
    =
    (1-\sigma)z+\sigma\epsilon,
    \qquad
    \epsilon\sim\mathcal{N}(0,\mathbf{I}),
    \label{eq:flow_interpolation}
\end{equation}
where $z$ denotes a clean video latent and $\epsilon$ is standard Gaussian noise.
Given the flow velocity \(v_{\theta}(x_{\sigma},\sigma)\) predicted by the video diffusion model, the corresponding clean latent estimate is
\begin{equation}
    \widehat{z}_{\theta}
    =
    x_{\sigma}
    -
    \sigma v_{\theta}(x_{\sigma},\sigma).
    \label{eq:clean_prediction}
\end{equation}
Replacing $\widehat{z}_{\theta}$ in~\Cref{eq:clean_prediction} with $z^{\mathrm{g}}$ and solving for the velocity yields the guidance velocity:
\begin{equation}
    v_{\mathrm{g}}(x_{\sigma},\sigma)
    =
    \frac{x_{\sigma}-z^{\mathrm{g}}}{\sigma},
    \label{eq:reference_velocity}
\end{equation}
which defines the flow velocity whose corresponding clean latent
estimate is $z^{\mathrm{g}}$.

We then steer the denoising trajectory toward the physical sequence by injecting the low-frequency residual between the guidance and predicted velocities.
Using the spatial low-pass operator $\mathcal{L}_{\rho}$ in~\Cref{eq:spatial_lowpass}, the modified velocity is given by
\begin{equation}
    \widetilde{v}_{\theta}
    =
    v_{\theta}
    +
    \mathcal{L}_{\rho}
    \left(
        v_{\mathrm{g}}-v_{\theta}
    \right),
    \label{eq:velocity_injection}
\end{equation}
where function arguments are omitted for brevity.
With this formulation, the low-frequency spatial components of the predicted velocity are replaced by those of the guidance velocity, while high-frequency components outside the mask remain unchanged.
We apply this velocity injection during the early denoising steps to establish global spatiotemporal structure, leaving the later steps unguided to refine fine visual details~\cite{yatim2024space}.

\begin{figure*}[t]
    \centering
    \includegraphics[width=\textwidth]{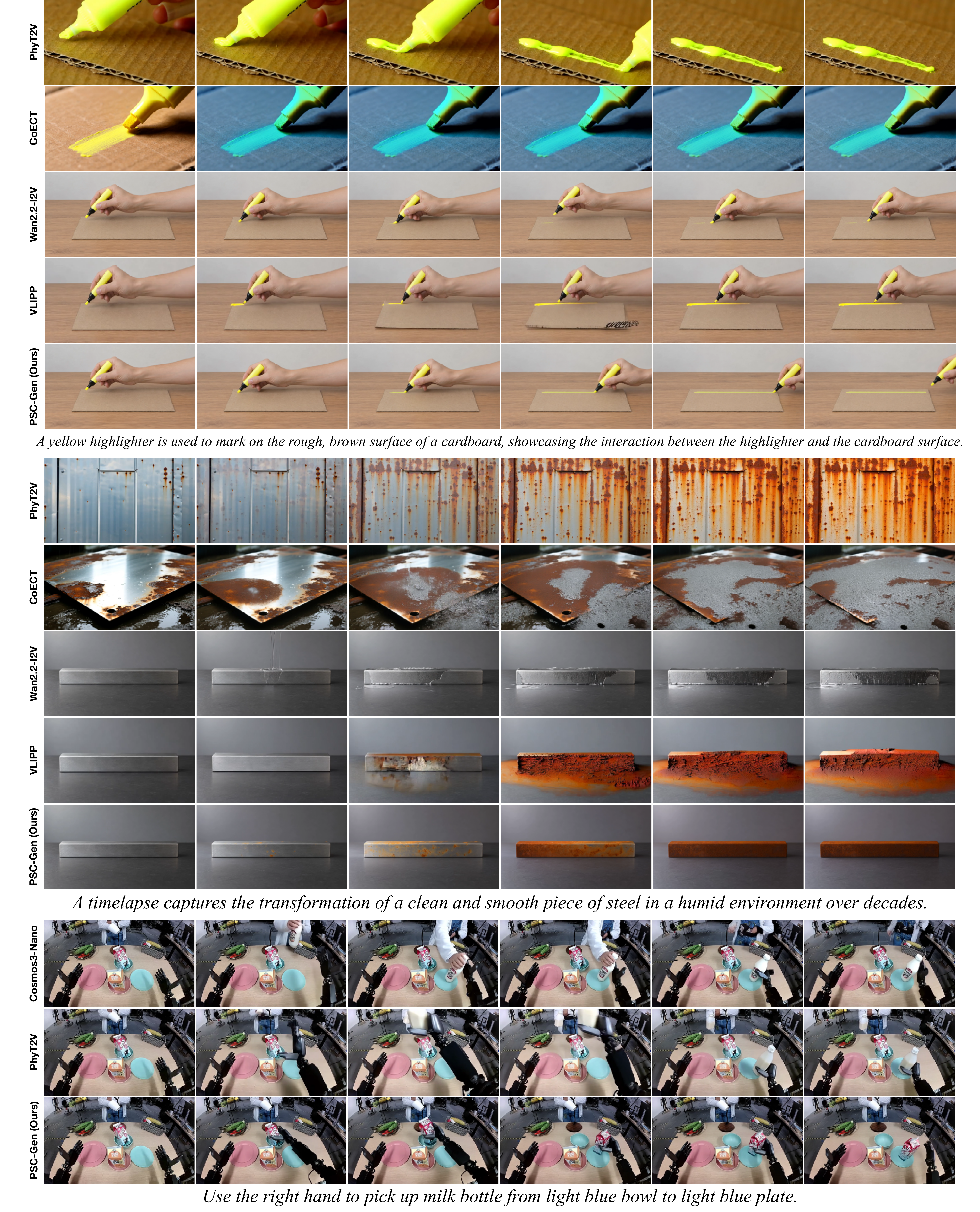}
    \vspace{-1.5em}
    \caption{\textbf{Qualitative comparison between PSC-Gen and baseline models across diverse physical scenarios.}}
    \label{fig:qualitative_result}
\end{figure*}

\section{Experiments}
\label{sec:experiment}

We evaluate PSC-Gen across a diverse suite of benchmarks~\cite{phygenbench,jang2025dreamgen,physicsiq} spanning general physical dynamics~\cite{phygenbench,physicsiq} and robotic manipulation~\cite{jang2025dreamgen}.
Our experiments assess both physical plausibility and prompt adherence, with DreamGenBench~\cite{jang2025dreamgen} specifically testing generalization to embodied world modeling.
In~\Cref{exp:experimental_setup}, we detail our experimental setup and evaluation protocol, and in~\Cref{exp:qualitative_results}, we present qualitative comparisons.
We then report quantitative evaluations in~\Cref{exp:quantitative_results}, followed by ablation studies validating our key design choices in~\Cref{exp:ablation_study}.

\subsection{Experimental Setup}
\label{exp:experimental_setup}

\paragraph{Baselines.}
To verify that our method can be seamlessly adopted into both general video models and world models, we use Wan2.2~\cite{wan22} and Cosmos3~\cite{nvidia2026cosmos3omnimodalworld} as our primary baseline backbones and apply PSC-Gen to each.
We further compare against a text-based approach, PhyT2V~\cite{xue2025phyt2vllmguidediterativeselfrefinement}, alongside two visual-proxy approaches, VLIPP~\cite{yang2025vlipp} and CoECT~\cite{wang2026chainCoECT}.
For a fair comparison, we adapt PhyT2V to match our backbones.
Because PhyT2V operates via iterative text refinement, we use Wan2.2-T2V on prompt-only PhyGenBench~\cite{phygenbench}, and Wan2.2-I2V and Cosmos3-Nano on image-conditioned Physics-IQ~\cite{physicsiq} and DreamGenBench~\cite{jang2025dreamgen}, respectively.
For CoECT, we retain its original planning procedure, Qwen-based image editing module~\cite{wu2025qwenimagetechnicalreport}, and CogVideoX~\cite{yang2024cogvideox} backbone in the main comparisons, as its diffusion-based keyframe conditioning does not directly generalize to flow-matching video backbones~\cite{wan22, nvidia2026cosmos3omnimodalworld, kong2025hunyuanvideosystematicframeworklarge}.
To further ensure a controlled comparison under identical foundation models, we also adapt CoECT to our exact backbone models in~\Cref{exp:ablation_study}.
Finally, because VLIPP requires a specifically fine-tuned motion-conditioned diffusion model~\cite{burgert2025gowiththeflow}, we retain its original pipeline and supply the initial image $I_0$ generated by our Physics Agent on prompt-only benchmark so that both methods start from identical visual states.

\paragraph{Evaluation Benchmarks.}
We evaluate PSC-Gen across two physical reasoning benchmarks~\cite{phygenbench,physicsiq} and one robotic manipulation benchmark~\cite{jang2025dreamgen}.
Specifically, we use PhyGenBench~\cite{phygenbench} and Physics-IQ~\cite{physicsiq} to assess our model's ability to generate physically plausible videos while faithfully following text prompts.
In addition, we adopt DreamGenBench~\cite{jang2025dreamgen} to demonstrate that PSC-Gen effectively generalizes to world model backbones such as Cosmos3~\cite{nvidia2026cosmos3omnimodalworld} for instruction-following, physically plausible robot manipulation synthesis.
We follow the official evaluation pipeline for each benchmark.
However, because the GPT-4o~\cite{openai2024gpt4ocard} judge frequently hallucinates from language priors when inspecting generated videos~\cite{li2025worldmodelbench, li2025videohallu, jang2025dreamgen}, we replace GPT-4o with a stronger VLM judge, Gemini-3.8-Flash~\cite{google2026gemini38flash}, for PhyGenBench and DreamGenBench following recent practice~\cite{gu2026phyworldbench, liao2024devil, lee2026bootstrapping}.
For completeness, evaluation scores using the default GPT-4o judge and full protocol details are provided in the supplementary material.

\paragraph{Implementation Details.}
We implement PSC-Gen on two primary video backbones: Wan2.2~\cite{wan22} and Cosmos3~\cite{nvidia2026cosmos3omnimodalworld}.
Specifically, we use Wan2.2-I2V-A14B for PhyGenBench and Physics-IQ, and Cosmos3-Nano for DreamGenBench.
For the Physics Agent, we employ Qwen3.8-27B~\cite{qwen38}, paired with GPT-Image-2.5~\cite{openai2026gptimage25} for keyframe synthesis and LTX-2.5~\cite{hacohen2025ltx25} for physical sequence interpolation.
For low-frequency velocity injection, we set the spatial frequency cutoff to $\rho = 0.15$ and apply guidance during the first 4 denoising steps, corresponding to $\sigma \ge 0.97$ for Wan2.2 and $\sigma \ge 0.99$ for Cosmos3.
The same planning and physical sequence synthesis pipeline is shared across both video backbones.
In our default configuration, we generate 5-second videos at 16 frames per second.

\subsection{Qualitative Results}
\label{exp:qualitative_results}

In~\Cref{fig:qualitative_result}, we present qualitative comparisons between PSC-Gen and baseline models across diverse scenarios.
Across all settings, PSC-Gen synthesizes physically plausible videos whereas the baseline models struggle to capture the intended physical outcomes.

The first scenario of~\Cref{fig:qualitative_result} clearly illustrates how each baseline struggles to generate physically plausible videos compared to our method.
Given the prompt to mark a rough cardboard surface using a yellow highlighter, PhyT2V renders a yellow streak that resembles thick wet paint rather than highlighter ink, as text alone struggles to convey fine material appearance.
Meanwhile, CoECT exhibits color-shifting artifacts caused by its keyframe conditioning during denoising.
Wan2.2-I2V also fails to produce a plausible interaction, showing the highlighter repeatedly tapping the cardboard and leaving only a faint mark.
Although VLIPP generates a bright line, its coarse bounding-box trajectory only specifies motion direction and misses physical contact between the tip and the surface.
In contrast, our PSC-Gen maintains contact between the highlighter and the cardboard while drawing a yellow line, demonstrating the effectiveness of our physically consistent visual conditioning.

The second scenario of~\Cref{fig:qualitative_result} and~\Cref{fig:teaser}(a) evaluate physical reasoning cases involving long-term oxidation of steel and air evacuation from a sealed aluminum can.
In both cases, Wan2.2-I2V fails to infer the implied physical outcomes from the prompt, leaving the steel bar unrusted and the can static throughout the video.
In the steel oxidation scenario, PhyT2V and CoECT depict rust over time, yet both violate the initial state constraint by showing rust already present in the first frame even though clean steel is specified.
Although VLIPP starts from a clean steel bar, it erroneously plans an expanding bounding-box trajectory for the steel, causing the metal bar itself to unnaturally swell alongside the rust.
By contrast, PSC-Gen accurately reasons about both physical processes, gradually spreading rust across an initially clean steel surface and collapsing the evacuated can according to the planned physical scenario.

\begin{table}[t]
    \centering
    \caption{\textbf{PhyGenBench Results.} Performance is evaluated across four physical domains and the overall average. Best results shown in bold, second best underlined.}
    \vspace{-0.5em}
    \label{tab:phygenbench_results}
    \footnotesize
    \setlength{\tabcolsep}{3.5pt}
    \resizebox{\columnwidth}{!}{%
    \begin{tabular}{@{}l*{5}{>{\centering\arraybackslash}p{1.1cm}}@{}}
        \toprule
        Method & Mechanics & Optics & Thermal & Material & Overall \\
        \midrule
        Wan2.2-I2V~\cite{wan22}
        & 0.550
        & 0.610
        & 0.556
        & 0.579
        & 0.577 \\
        PhyT2V~\cite{xue2025phyt2vllmguidediterativeselfrefinement}
        & \underline{0.625}
        & \underline{0.620}
        & \underline{0.633}
        & \underline{0.683}
        & \underline{0.640} \\
        VLIPP~\cite{yang2025vlipp}
        & 0.596
        & 0.570
        & 0.550
        & 0.583
        & 0.576 \\
        CoECT~\cite{wang2026chainCoECT}
        & 0.538
        & 0.557
        & 0.528
        & 0.504
        & 0.533 \\
        \midrule
        \textbf{PSC-Gen (Ours)}
        & \textbf{0.688}
        & \textbf{0.647}
        & \textbf{0.683}
        & \textbf{0.742}
        & \textbf{0.688} \\
        \bottomrule
    \end{tabular}%
    }
\end{table}

\begin{table}[t]
\vspace{-0.5em}
\centering
\caption{\textbf{DreamGenBench Results.} Performance is evaluated across Object, Behavior, and Environment categories and the overall average. Instruction Following (IF) measures prompt adherence and Physics Alignment (PA) measures physical plausibility. Best results shown in bold, second best underlined.}
\vspace{-0.5em}
\label{tab:dreamgenbench_results}
\footnotesize
\setlength{\tabcolsep}{3pt}
\resizebox{\columnwidth}{!}{%
\begin{tabular}{@{}l*{8}{c}@{}}
\toprule
& \multicolumn{2}{c}{Object}
& \multicolumn{2}{c}{Behavior}
& \multicolumn{2}{c}{Environment}
& \multicolumn{2}{c}{Overall} \\
\cmidrule(lr){2-3}\cmidrule(lr){4-5}
\cmidrule(lr){6-7}\cmidrule(lr){8-9}
Method & IF & PA
& IF & PA
& IF & PA
& IF & PA \\
\midrule
Cosmos3-Nano~\cite{nvidia2026cosmos3omnimodalworld}
& \underline{54.00} & 6.81
& \underline{61.70} & \underline{6.93}
& \underline{65.52} & \underline{24.29}
& \underline{59.52} & 10.88 \\
PhyT2V~\cite{xue2025phyt2vllmguidediterativeselfrefinement}
& \underline{54.00} & \underline{9.11}
& 51.06 & 5.63
& 58.62 & 23.77
& 53.97 & \underline{11.19} \\
\midrule
\textbf{PSC-Gen (Ours)}
& \textbf{78.00} & \textbf{12.66}
& \textbf{80.85} & \textbf{8.41}
& \textbf{72.41} & \textbf{28.08}
& \textbf{77.78} & \textbf{14.62} \\
\bottomrule
\end{tabular}%
}
\vspace{-2em}
\end{table}

\Cref{fig:teaser}(b) and the third scenario of~\Cref{fig:qualitative_result} evaluate robotic manipulation synthesis.
In~\Cref{fig:teaser}(b), the Cosmos3-Nano baseline struggles to follow the spatial goal and repeatedly scoops inside the left bowl, whereas PSC-Gen picks up the spoon and executes a clear scooping motion from the left bowl to the right bowl.
Similarly, in the pick-and-place scenario of~\Cref{fig:qualitative_result}, both Cosmos3-Nano and PhyT2V fail to locate the target object and instead hallucinate a new milk bottle emerging from the white cup in the person's hand.
By contrast, PSC-Gen accurately identifies the milk bottle inside the light blue bowl and places it onto the light blue plate.
Together, these results confirm that PSC-Gen generalizes effectively across both general video models (Wan2.2) and world model backbones (Cosmos3), delivering coherent physical planning and instruction-following synthesis across diverse scenarios.
Additional qualitative comparisons are provided in the supplementary material.

\begin{table}[t]
    \centering
    \caption{\textbf{Physics-IQ Results.} Performance by domain and overall, excluding lit-candle cases. Higher scores are better. Best and second-best results are shown in bold and underlined, respectively.}
    \label{tab:physicsiq_results}
    \vspace{-0.5em}
    \scriptsize
    \setlength{\tabcolsep}{3pt}
    \resizebox{\columnwidth}{!}{%
    \begin{tabular}{@{}l*{6}{>{\centering\arraybackslash}p{0.8cm}}@{}}
        \toprule
        Method & Solid & Fluid & Optics & Thermo & Magnet & Overall \\
        \midrule
        Wan2.2-I2V~\cite{wan22}
        & 43.05
        & \underline{41.21}
        & \textbf{35.55}
        & 21.70
        & 15.76
        & \underline{39.04} \\
        PhyT2V~\cite{xue2025phyt2vllmguidediterativeselfrefinement}
        & \underline{43.78}
        & 34.65
        & 33.60
        & \underline{28.95}
        & \underline{21.86}
        & 38.82 \\
        VLIPP~\cite{yang2025vlipp}
        & 27.14
        & 31.68
        & 16.46
        & 17.54
        & 6.63
        & 24.55 \\
        CoECT~\cite{wang2026chainCoECT}
        & 22.98
        & 20.84
        & 8.62
        & 15.58
        & 20.66
        & 19.71 \\
        \midrule
        \textbf{PSC-Gen (Ours)}
        & \textbf{45.00}
        & \textbf{42.73}
        & \underline{34.81}
        & \textbf{33.13}
        & \textbf{26.76}
        & \textbf{41.02} \\
        \bottomrule
    \end{tabular}%
    }
\vspace{-0.5em}
\end{table}

\begin{figure}[t]
    \centering
    \includegraphics[width=\columnwidth]{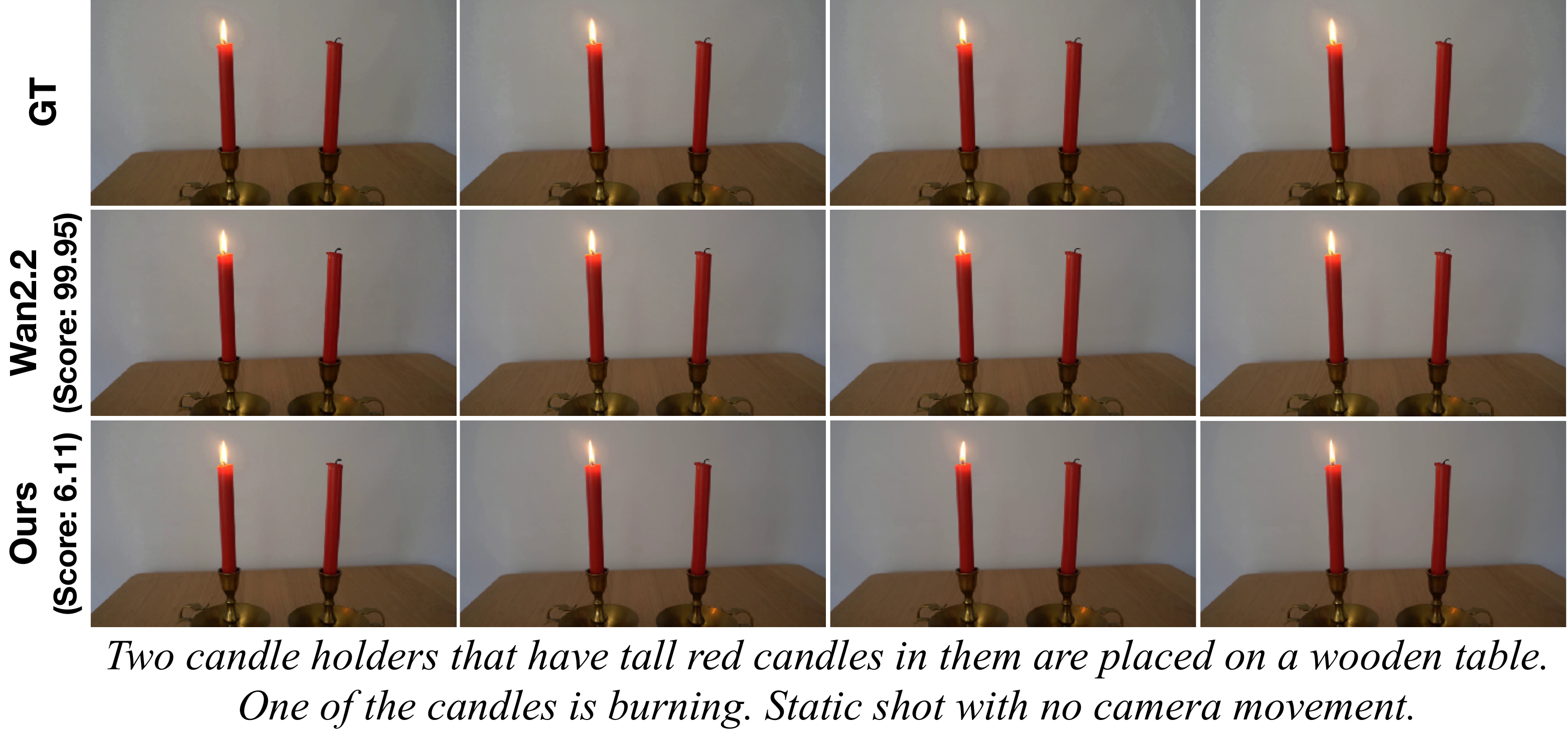}
    \vspace{-1.5em}
    \caption{Comparison of ground truth, Wan2.2, and PSC-Gen (Ours) on a lit-candle example from Physics-IQ. Wan2.2 receives a much higher score, although the only apparent difference is natural flame motion.}
    \label{fig:physics_iq}
    \vspace{-1.5em}
\end{figure}

\begin{figure*}[t]
    \centering
    \includegraphics[width=\textwidth]{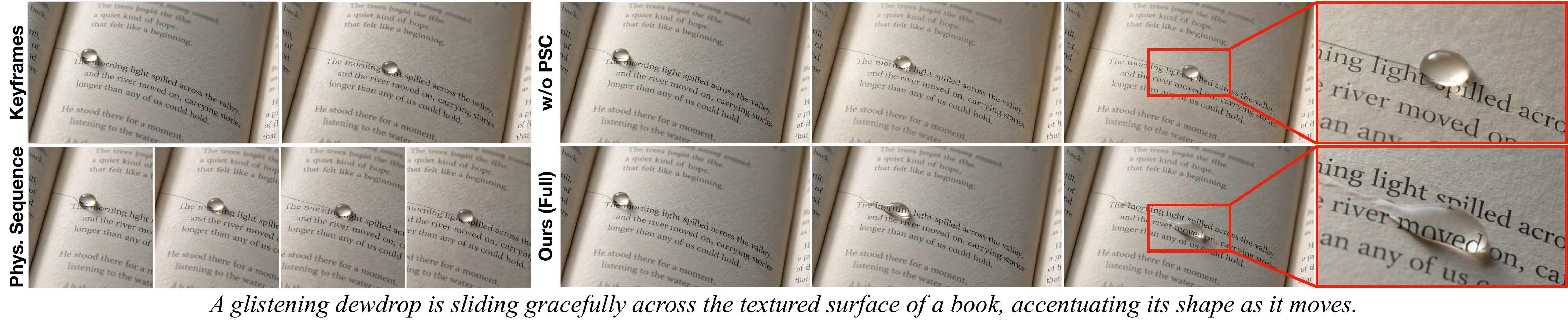}
    \vspace{-1.2em}
    \caption{\textbf{Ablation Study on Physical Sequence Conditioning.} The top row shows the variant without Physical Sequence Conditioning, and the bottom row shows our full pipeline. While the ablated variant simply reconstructs and interpolates the static keyframes, our full pipeline achieves more natural, physically plausible motion guided by the physical sequence.}
    \label{fig:ablation}
    \vspace{-0.7em}
\end{figure*}

\subsection{Quantitative Results}
\label{exp:quantitative_results}

We evaluate our method across multiple benchmarks for physical reasoning and robotic manipulation video generation.
We first evaluate physical reasoning and plausible motion synthesis on PhyGenBench~\cite{phygenbench}.
As reported in~\Cref{tab:phygenbench_results}, PSC-Gen achieves the highest score across every physical domain among all baselines.
Specifically, PSC-Gen improves the overall score over the Wan2.2-I2V backbone by 19\% and surpasses PhyT2V by 7.5\%, even though PhyT2V requires $3\times$ iterative video generation per prompt.

To assess generalization to robotic manipulation, we evaluate PSC-Gen on DreamGenBench.
\Cref{tab:dreamgenbench_results} shows that PSC-Gen outperforms all baselines across every category, achieving 77.78 on Instruction Following (IF) and 14.62 on Physics Alignment (PA).
Compared to the Cosmos3-Nano base model, PSC-Gen improves overall IF by 31\% and PA by 34\%, while also surpassing PhyT2V on PA by 31\%.

We further evaluate real-world physical dynamics on Physics-IQ against ground-truth recordings.
As shown in~\Cref{tab:physicsiq_results}, PSC-Gen ranks best on Solid, Fluid, Thermo, and Magnetism, second on Optics, and highest overall, outperforming Wan2.2-I2V by 5.1\%.
\Cref{fig:physics_iq} illustrates a metric limitation in the lit-candle scenario: although PSC-Gen synthesizes a naturally flickering flame, Physics-IQ's pixel metrics penalize this motion against the static ground-truth flame, scoring PSC-Gen at 6.11 versus 99.95 for Wan2.2-I2V's static output.
Consistent with recent findings~\cite{radsch2026physicsiqverified} that reference-based metrics can penalize plausible continuations, we report scores excluding the 3 lit-candle cases out of 198 videos in~\Cref{tab:physicsiq_results}, with full unfiltered scores in the supplementary material.
Overall, these results confirm that PSC-Gen consistently improves physical plausibility and instruction adherence across general physics and robotic manipulation benchmarks.

\begin{table}[t]
    \centering
    \caption{\textbf{Ablation Study on PhyGenBench.} Comparison of Wan2.2-I2V, the CoECT variant, three ablated variants, and our full pipeline. Best in bold, second best underlined.}
    \vspace{-0.5em}
    \label{tab:ablation}
    \footnotesize
    \setlength{\tabcolsep}{2pt}
    \renewcommand{\arraystretch}{1.12}
    \resizebox{\columnwidth}{!}{%
    \begin{tabular}{@{}l*{5}{c}@{}}
        \toprule
        Method & Mechanics & Optics & Thermal & Material & Overall \\
        \midrule
        Wan2.2-I2V~\cite{wan22} (Base)
        & 0.542
        & 0.627
        & 0.633
        & 0.569
        & 0.587 \\
        CoECT~\cite{wang2026chainCoECT} Variant
        & 0.617
        & 0.653
        & 0.611
        & \textbf{0.708}
        & 0.652 \\
        w/o Agentic Visual Planning
        & \textbf{0.688}
        & 0.647
        & \underline{0.700}
        & 0.583
        & 0.650 \\
        w/o Physical Sequence Conditioning
        & 0.625
        & \underline{0.657}
        & 0.667
        & \underline{0.681}
        & 0.653 \\
        w/o Low-Frequency Filtering
        & \underline{0.677}
        & \underline{0.657}
        & 0.633
        & 0.667
        & \underline{0.663} \\
        \midrule
        \textbf{PSC-Gen (Full)}
        & \underline{0.677}
        & \textbf{0.706}
        & \textbf{0.733}
        & \textbf{0.708}
        & \textbf{0.700} \\
        \bottomrule
    \end{tabular}%
    }
    \vspace{-2.0em}
\end{table}

\subsection{Ablation Study}
\label{exp:ablation_study}

We conduct ablation studies on PhyGenBench~\cite{phygenbench}, using a randomly sampled 50-prompt subset across all four physical domains, to validate our core design choices.
To directly isolate the benefit of our framework over prior keyframe-based conditioning under identical model capacity, we also evaluate a CoECT~\cite{wang2026chainCoECT} variant adapted to use the exact same language model, image generator, and video backbone as ours.
As shown in~\Cref{tab:ablation}, although this CoECT variant improves upon the Wan2.2-I2V~\cite{wan22} base model, our full PSC-Gen pipeline still outperforms it by 7.4\%.
Among our ablated variants, replacing Agentic Visual Planning with a simple general-purpose VLM planner reduces the overall score by 7.1\%, which confirms that domain-grounded agentic physical reasoning is essential for general physical plausibility.
Removing Physical Sequence Conditioning lowers the overall score by 6.7\%, demonstrating the necessity of dense spatiotemporal sequence guidance over sparse static keyframe conditioning.
Finally, removing spatial low-pass filtering decreases the overall score by 5.3\%, confirming that isolating low-frequency components is critical for transferring coarse physical dynamics without propagating high-frequency visual artifacts.

We also compare the qualitative results of the variant without Physical Sequence Conditioning and our full pipeline in~\Cref{fig:ablation}.
In the top row, the conditioned keyframes on the left show dewdrops with rigid, nearly identical shapes, and the generated frames on the right simply reconstruct and interpolate them into unnaturally uniform droplets sliding across the book.
In the bottom row, the left four frames from the interpolated physical sequence establish a coherent motion path but still inherit the rigid dewdrop appearance from the keyframes.
In contrast, the right three images from our full pipeline show the dewdrop flowing naturally to the right along the guided trajectory while the primary diffusion model synthesizes realistic, physically plausible details.
These results confirm that PSC-Gen effectively injects coarse physical dynamics into the denoising process while preserving high-fidelity visual synthesis.

\section{Conclusion}
\label{sec:conclusion}

We present PSC-Gen, a training-free framework for physically plausible video generation faithful to input prompts.
By iteratively planning and verifying each visual state through Agentic Visual Planning, PSC-Gen ensures physical consistency across the synthesized visual conditions.
Physical Sequence Conditioning further interpolates these states into a sequence and injects their low-frequency physical dynamics to effectively guide the denoising process toward physically plausible videos.
This design equips off-the-shelf video generators with explicit physical reasoning capabilities without additional data collection or extensive model training.
Comprehensive experiments across general physics and robotic manipulation benchmarks confirm that PSC-Gen enables pretrained video models to achieve superior physical dynamics and instruction adherence.
Qualitative comparisons further illustrate that our approach accurately captures complex physical phenomena across diverse scenarios from simple text instructions.

\section*{Acknowledgement}
{\emergencystretch=1em
This research was partially supported by the AMD University Program (AUP).\par}
{
    \small
    \bibliographystyle{ieeenat_fullname}
    \bibliography{main}
}


\end{document}